\documentclass[journal]{IEEEtran}
\ifCLASSINFOpdf
\else
\fi
\usepackage{indentfirst}
\usepackage{graphicx}
\usepackage[justification=centering]{caption}
\usepackage{subfigure}
\usepackage{amsmath}
\usepackage{tabularx}
\usepackage{multirow}
\usepackage{siunitx}
\usepackage{color}
\usepackage[numbers]{natbib}
\usepackage[colorlinks,linkcolor=blue]{hyperref}
\usepackage{amssymb}

\begin{document}
%
\title{A Single Frame and Multi-Frame Joint Network for 360-degree Panorama Video Super-Resolution}
%
%

\author{Hongying~Liu,~\IEEEmembership{Member,~IEEE,}
            Zhubo~Ruan,
            Chaowei~Fang,~\IEEEmembership{Member,~IEEE,}
            Peng~Zhao,\\
            Fanhua~Shang,~\IEEEmembership{Senior Member,~IEEE,}
            Yuanyuan~Liu,~\IEEEmembership{Member,~IEEE,}
            and~Lijun~Wang
\thanks{H.\ Liu, Z.\ Ruan, C.\ Fang, P.\ Zhao, F.\ Shang (Corresponding author), and Y.\ Liu are with the Key Laboratory of Intelligent Perception and Image Understanding of Ministry of Education, School of Artificial Intelligence, Xidian University, Xi'an, China. E-mails: \{hyliu, cwfang, fhshang, yyliu\}@xidian.edu.cn, zboruan@163.com.}
\thanks{L.\ Wang is with the Information School, North China University of Technology, Beijing, China. E-mail: ljwang@outlook.com.}
\thanks{Manuscript received August 21, 2020.}}

\maketitle

\begin{abstract}
Spherical videos, also known as \ang{360} (panorama) videos, can be viewed with various virtual reality devices such as computers and head-mounted displays. They attract large amount of interest since awesome immersion can be experienced when watching spherical videos. However, capturing, storing and transmitting high-resolution spherical videos are extremely expensive.
In this paper, we propose a novel single frame and multi-frame joint network (SMFN) for recovering high-resolution spherical videos from low-resolution inputs. To take advantage of pixel-level inter-frame consistency, deformable convolutions are used to eliminate the motion difference between feature maps of the target frame and its neighboring frames. A mixed attention mechanism is devised to enhance the feature representation capability. The dual learning strategy is exerted to constrain the space of solution so that a better solution can be found. A novel loss function based on the weighted mean square error is proposed to emphasize on the super-resolution of the equatorial regions. This is the first attempt to settle the super-resolution of spherical videos, and we collect a novel dataset from the Internet, MiG Panorama Video, which includes 204 videos. Experimental results on 4 representative video clips demonstrate the efficacy of the proposed method. The dataset and code are available at \href{https://github.com/lovepiano/SMFN_For_360VSR}{https://github.com/lovepiano/SMFN\_For\_360VSR}.
\end{abstract}

\begin{IEEEkeywords}
\ang{360} panorama videos, virtual reality, video super-resolution, weighted mean square error
\end{IEEEkeywords}


%
\IEEEpeerreviewmaketitle

\section{Introduction}
\IEEEPARstart{W}{ixel} recent development of virtual reality (VR) technology and the 5th generation of mobile networks, \ang{360} panorama videos become increasingly important for immersive technologies. This video format can provide highly immersive experiences for audiences. Existing VR display devices for spherical or panorama videos, such as head-mounted displays, allows for up to 6 degrees of freedom in the interaction with the captured virtual environment. Panorama videos are captured with \ang{360} cameras and stored in two-dimensional planar format, such as equirectangular projection (ERP) and cubemap projection. During playback, the panorama videos can be projected to both normal flat displays and spherical displays. Recently, panorama videos have been applied in a variety of fields, such as entertainment \cite{panorama1}, education, health-care, communication \cite{panorama2}, and advertising.

However, this video format brings great challenges because it needs very high resolution to cover the entire \ang{360} space to make sure that the captured videos are visually satisfactory. A part of a given panorama video, called the viewport \cite{panorama3} can be exploited by a VR display. The resolution of the viewport is relatively low, thus the visual effect is greatly reduced. Thus, high resolution videos are paramount to guarantee the high quality of the viewing experience in VR systems based on spherical videos. For example, in terms of recent studies such as \cite{panorama5}, the resolution of the captured panorama videos should be 21,600$\times$10,800, so that the audiences can obtain high immersion sense in the captured virtual environment. However, currently both of consumer capturing systems and network bandwidths are difficult to process and transmit such large-scale video in real time. An effective way to solve the above problem is to capture low-resolution videos at first, and then super-resolve them into high-resolution videos. In this paper, we propose an effective method based on deep learning to settle the problem of \ang{360} video super-resolution.

With the advent of deep learning, super-resolution algorithms for images/videos have evolved from traditional interpolation methods to predictive or generative models based on convolutional neural networks (CNNs). Due to the strong feature learning capability of CNNs, great success has been achieved for image/video super-resolution (SR). For generic videos, high-quality reconstruction performance can be achieved via recent high-capacity algorithms based on deep learning, such as the recurrent back-projection network (RBPN) \cite{RBPN} and the enhanced deformable video restoration (EDVR) \cite{EDVR}. However, there is no published algorithms using deep learning to solve the \ang{360} panorama video super-resolution task. Therefore, this is the first research work for \ang{360} panorama video super-resolution. Specially, in the paper, we process the panorama videos in ERP format.

\textbf{Our contributions:} In this paper, we propose the first deep learning method for super-resolution of \ang{360} panorama videos. In particular, we design a novel single frame and multi-frame joint panorama video super-resolution model with dual learning. It takes low-resolution panorama videos as input and then outputs the corresponding high-resolution counterparts.
For \ang{360} panorama videos, important contents are usually displayed in equatorial regions, so we propose a weighted loss that aims to increase the weights of the equatorial regions while decreasing the weights of the polar regions. This loss function makes our approach be capable of focusing on the most important regions and thus enhance the quality of panorama videos. To the best of our knowledge, there is no work using deep learning for the super-resolution of \ang{360} panorama videos, and there is no publicly available dataset. For this reason, we construct a dataset, the MiG panorama video dataset, by collecting a large number of panorama videos from the Internet, and edit 204 panorama videos in total. Each video includes 100 frames with resolution ranging from 4,096$\times$2,048 to 1,440$\times$720. The captured scenes vary from outdoor to indoor, and from day to night. They will be made publicly available.

Our main contributions are summarized in three-fold.

$\bullet$ We explore the super-resolution of \ang{360} panorama videos using deep learning for the first time, and propose a novel panorama video super-resolution model.

$\bullet$ We design a single frame and multi-frame joint network and provide a weighted loss function to make the network pay more attention to restoration of the equatorial regions.

$\bullet$ We build the first dataset for the super-resolution of \ang{360} panorama videos. We hope that our new insight could potentially deepen the understanding to the studies of panorama video super-resolution.

The rest of this paper is organized as follows. The related work for super-resolution is described in Section II. Section III presents the proposed method. In Section IV, we show the exciting experimental results of our method. Finally, conclusions and future work are presented in Section V.

\section{Related Work}
In this section, we briefly introduce the development of super-resolution techniques based on deep learning in recent years, including the single image super-resolution techniques and video super-resolution techniques.

\subsection{Single Image Super-Resolution}
Single image super-resolution methods aim at learning the mapping between low-resolution images and high-resolution images, they take a single low-resolution image as input and output a corresponding super-resolved image \cite{wang2020deep}. As the first work based on deep learning, the super-resolution convolutional neural network (SRCNN) \cite{SRCNN} started the era of deep learning in the field of image super-resolution. After that, more and more super-resolution algorithms based on deep learning have been proposed. For instance, to reduce the computation cost caused by using interpolated HR images as input, subsequent algorithms such as \cite{FSRCNN,ESPCN,EDSR,RCAN} use low-resolution images as input and then upsample the feature maps at the end of the network. Inspired by ResNet \cite{ResNet} and the idea that deeper networks can extract stronger and more complicate high-level/low-level features, several methods such as \cite{EDSR,RCAN,autoencoder,DBPN,RDN} adopt deeper network structures and meanwhile use residual structures to avoid vanishing gradients. Moreover, non-local block \cite{NLN,res_nonlocal,nonlocal_rcnn}, attention mechanism \cite{RCAN,CBAM,SAN}, feedback mechanism \cite{FB,SRFBN}, and dual learning methods \cite{dual_1,dual_2,dual_3,dual_4,DRN} are adopted to further improve the performance of image super-resolution.

All the above algorithms use generic images (i.e., captured with a fixed viewpoint) for training, while only few related works \cite{sr360,vrsrcnn2018,vrw2020,OurSurvey} were proposed for super-resolution of \ang{360} panorama images.

\subsection{Video Super-Resolution}
Video super-resolution usually takes multiple neighboring images as input and then output the super-resolution results of target frames. In recent years, with the rapid development of deep learning, many video super-resolution methods have been proposed. We categorize them into two main categorises: aligned and non-aligned methods according to whether the video frames are aligned in the method, as discussed in \cite{OurSurvey}. Please refer to the video super-resolution survey \cite{OurSurvey} for more methods and details. There are some aligned methods such as VSRnet \cite{VSRnet}, the video efficient sub-pixel convolutional network (VESPCN) \cite{VESPCN}, the task-oriented flow (TOFlow) \cite{TOFlow}, the frame recurrent video super-resolution (FRVSR) \cite{FRVSR}, and the recurrent back-projection network (RBPN) \cite{RBPN}. They exploit the inter-frame dependency through explicitly aligning different frames. The non-aligned methods mainly include VSRResNet \cite{VSRResNet}, the dynamic upsampling filters (DUF) \cite{DUF}, the fast spatio-temporal residual network (FSTRN) \cite{FSTRN}, the frame and feature-context video super-resolution (FFCVSR) \cite{FFCVSR}, and the 3D super-resolution network (3DSRNet) \cite{3DSRnet}. The relationship across frames is implicitly learned by the network itself. All the video super-resolution algorithms mainly adopt generic videos as training sets. For \ang{360} panorama videos, to the best of our knowledge, there is no research work of deep learning techniques for video super-resolution.

\section{A New Single Frame and Multi-Frame Joint Network}
In this section, we propose a novel single frame and multi-frame joint network (SMFN) for \ang{360} panorama video super-resolution. The entire network architecture of our SMFN method is firstly depicted, and then the details of each individual module are presented.

\begin{figure*}[ht]
\centering
\includegraphics[scale=0.6]{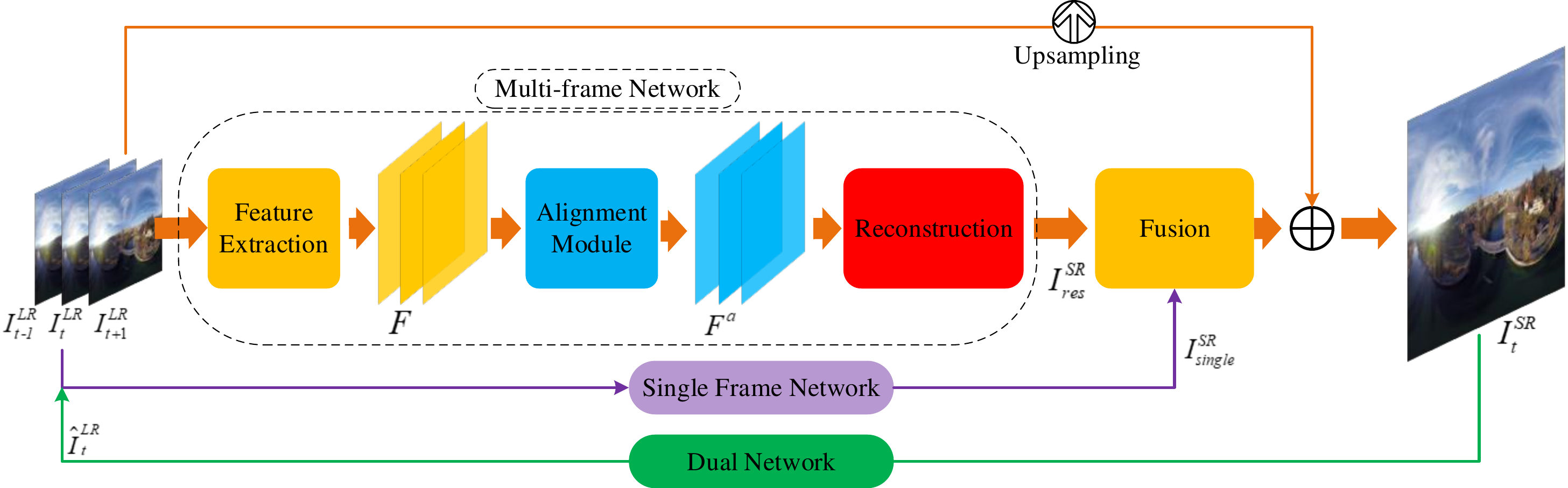}
\caption{The network architecture of our SMFN method, which is mainly composed of a multi-frame network (including feature extraction, alignment, and reconstruction modules), a single frame network, a fusion module and a dual network. Note that the dual network is only used in the training stage in order to constrain the solution space, which is beneficial for preventing the overfitting problem. The `upsampling' denotes the bilinear interpolation operation.}
\label{fw}
\end{figure*}

\subsection{Our Network Architecture}
The framework of our SMFN method is composed of a single frame super-resolution network, a multi-frame super-resolution network, a dual network, and a fusion module. The motivation of setting up this framework is to combine the advantages of both single frame and multi-frame super-resolution methods, which are good at recovering spatial information and exploring temporal information, respectively. The dual network is used to constrain the solution space. The architecture of the proposed SMFN is shown in Fig.\ \ref{fw}.

The goal of our network is to output the high-resolution counterpart of the target frame with scale factor $s$, given $2N+1$ ($N$ denotes the size of temporal window) low-resolution video frames $I_{t-N:t+N}^{LR}$ ($I_t^{LR}$ denotes the low-resolution target frame). As shown in Fig.\ \ref{fw}, firstly, the feature extraction module takes the target frame and its neighboring frames as inputs to produce the corresponding feature maps. Then the feature maps of the neighboring frames are registered with the feature maps of the target frame via the alignment module. The aligned features are fed into the reconstruction module to obtain the high-resolution image. The single frame network directly performs the single image super-resolution for the target frame. In order to further enhance the quality of the recovered videos, we also design a fusion module to post-process the results of the reconstruction module and the single frame network.
Alike in \cite{DRVSR,D3Dnet}, our proposed model only learns a residual map between the expected HR result and  the bi-linearly interpolated image of the LR input.
 The final super-resolution result is obtained by adding the upsampled LR target frame to the output of the network. For the dual network, it is only applied in the training phase to transfer the SR image back into the LR space. We use a regularization loss to make the output of the dual network be consistent with the original LR input. The details of each module are described in the following subsections.

\subsection{Single Frame Super-Resolution}

The purpose of using single frame image super-resolution is to take its advantage of recovering spatial information. In our proposed method, the single frame super-resolution module is constructed with multiple convolutional layers, and each convolution layer is followed by a ReLU activation layer. It directly takes the LR target frame as input and produce a preliminary SR image $I_{single}^{SR}$. The process of single frame super-resolution is formulated as follows:
\begin{equation}
I_{single}^{SR} = H_{single}(I_t^{LR}).
\end{equation}
where $H_{single} (\cdot)$ denotes the calculations of the single frame super-resolution network.

\subsection{Multi-Frame Super-Resolution}
The multi-frame network is a primary network for video super-resolution in our SMFN architecture, which utilizes the multiple input frames for feature learning and information recovery. It mainly includes feature extraction, alignment, reconstruction and fusion modules, which are illustrated below in detail.

\subsubsection{Shallow Feature Extraction}
The feature extraction module takes the consecutive LR frames $I_{t-N:t+N}^{LR}$ as input and yield a feature map for each input frame. This module is built upon residual blocks, which is composed of two convolutional layers and a ReLU activation function following the first convolutional layer. The entire process of the feature extraction module can be formulated as follows:
\begin{equation}
F = H_{FE}(I_{t-N:t+N}^{LR})
\end{equation}
where $H_{FE}(\cdot)$ denotes the feature extraction module, and $F$ represents the extracted feature map.

\subsubsection{Alignment Module}
We adopt the deformable convolutional network as in \cite{DCN,DCNV2} to perform the alignment operation between frames. A popular method for aligning different frames is to warp the source frame to the target frame with optical flow fields. However, the estimation of optical flow fields is very expensive and challenging. Deformable convolution is famous for modelling geometric transformations with a deformed sampling grid. A normal convolution can be easily modified into a deformable convolution via learning additional offsets for every position in the sampling grid. Here we use it to achieve inter-frame registration considering its efficiency.

Generally, a convolution operation with a regular grid is defined as follows:
\begin{equation}
F'(x_0)= \sum_{i=1}^{K}w_i{\cdot}F(x_0+p_i)
\end{equation}
where $K$ is equal to the square of kernel size. $w_i$ denotes the weight of convolution kernel at the $i$-th location. $F$ is the feature map. $x_0$ denotes the position in feature map $F'$. $p_i$ denotes the pre-specified offset for the $i$-th location. For instance, for a 3$\times$3 convolution kernel, $K$ is equal to 9 and $p_i \in \{{(-1,-1),(-1,0),...,(0,1),(1,1)}\}$.

Different from the general convolution, the deformable convolution uses dynamic offsets. Additional offsets are learned from the input feature maps to adjust the pre-specified offsets.
In this work, we take target features and neighboring features as the input of deformable convolution module to make it learn the offsets between target features and neighboring features, and then the neighboring features perform the convolution operation via learned offsets to achieve the alignment goal. The process can be formulated as follows:
\begin{equation}
F_{t+j}^a (x_0) = \sum_{j=1}^{K}w_j{\cdot}F_{t+j}(x_0+p_j+{\triangle}p_j)
\end{equation}
where $F_{t+j}^a$ denotes the feature maps aligned from $t+j$ to $t$, $F_{t+j}$ denotes the feature maps of frame $t+j$, and ${\triangle}p_j$ is learned offsets computed with the other convolutional layer, which takes the concatenation of $F_{t}$ and $F_{t+j}$ as input.

\subsubsection{Reconstruction Module}
The proposed reconstruction module consists of three parts: deep feature extraction and fusion, mixed attention, and an upscale module.

\begin{figure}[h]
\centering
\includegraphics[scale=0.7]{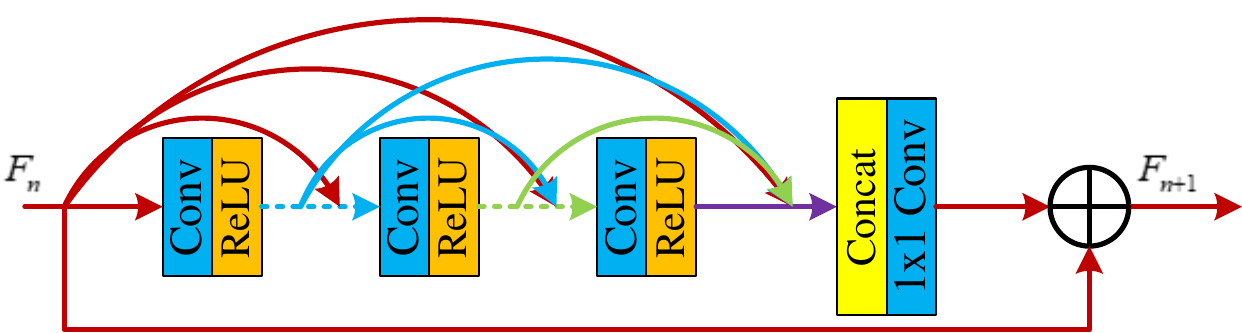}
\caption{The residual dense block in our reconstruction module.}
\label{RDB}
\end{figure}

\textbf{Deep feature extraction and fusion}:
We design a residual dense block (RDB) as our basic block for deep feature extraction and fusion, whose architecture is shown in Fig.\ \ref{RDB}.  In order to better fuse with the aligned features and obtain finer features, we adopt the this module for finer feature extraction and fusion as in \cite{resnet2}. In our method, each residual dense block consists of five 3$\times$3 convolutional layer followed by a ReLU activation function respectively and a 1$\times$1 transition layer without activation function. The residual learning can be formulated as follows:
\begin{equation}
F_{n+1} = F_n + H_{res}(F_n)
\end{equation}
where $H_{res}(F_n)$ denotes the learned residual, $F_n$ represents the input feature map of a residual block, and $F_{n+1}$ is the output feature maps.


\textbf{Mixed attention}:
We adopts the attention mechanism to further enhance the representative ability of the proposed network. Specifically, we combine the channel attention (CA) \cite{RCAN} and space attention (SA) \cite{dualattention}, which is called the mixed attention in the paper. The CA module aims at filtering out the redundant information among channels and highlighting important information. Likewise, the SA module aims at filtering out the redundant information in space and focusing on important regions. In our method, the input features are separately fed into CA module and SA module to obtain the corresponding channel and space attention maps. Next, we add the CA maps with the SA maps, resulting in mixed attention maps. Then the final output is the multiplication of the mixed attention maps and the initial input features.

\textbf{Upscale module}:
After extracting and fusing features in the LR space, a $3\times3$ convolutional layer with $s^2C$ output channels is adopted to produce $s^2C$ feature channels for the following sub-pixel convolutional layer. The sub-pixel convolutional layer converts the LR images with size of $H{\times}W{\times}s^2C$ to the corresponding HR images with size of $sH{\times}sW{\times}C$. Then a $3\times3$ convolutional layer is used to reconstruct the SR residual image $I_{res}^{SR}$. Next, the reconstructed image is fed into the fusion module to perform the fusion operation.

\subsection{Fusion Module}
 We devise this module to further enhance the performance of our network for video super-resolution, which fuse the spatial features from the single frame network with that of the multi-frame network. The fusion module consists of three convolutional layers with kernel size of $3\times3$. It takes the output of reconstruction module and single frame network as input to produce the fused output. The output of this module $I_{fus}^{SR}$ can be formulated as follows:
\begin{equation}
I_{fus}^{SR} = H_{fus}(I_{res}^{SR},I_{single}^{SR})
\end{equation}
where $H_{fus}(\cdot)$ denotes the operation of fusion module.

After the fusion module, the final output of our proposed method can be obtained by adding $I_{fus}^{SR}$ with the bilinearly upsampled LR target frame. The output $I_t^{SR}$ is formulated as
\begin{equation}
I_t^{SR} = I_{fus}^{SR} + bilinear(I_t^{LR}).
\end{equation}
where $bilinear(\cdot)$ denotes the bilinear interpolation algorithm.

\begin{figure}[ht]
\centering
\includegraphics[scale=0.65]{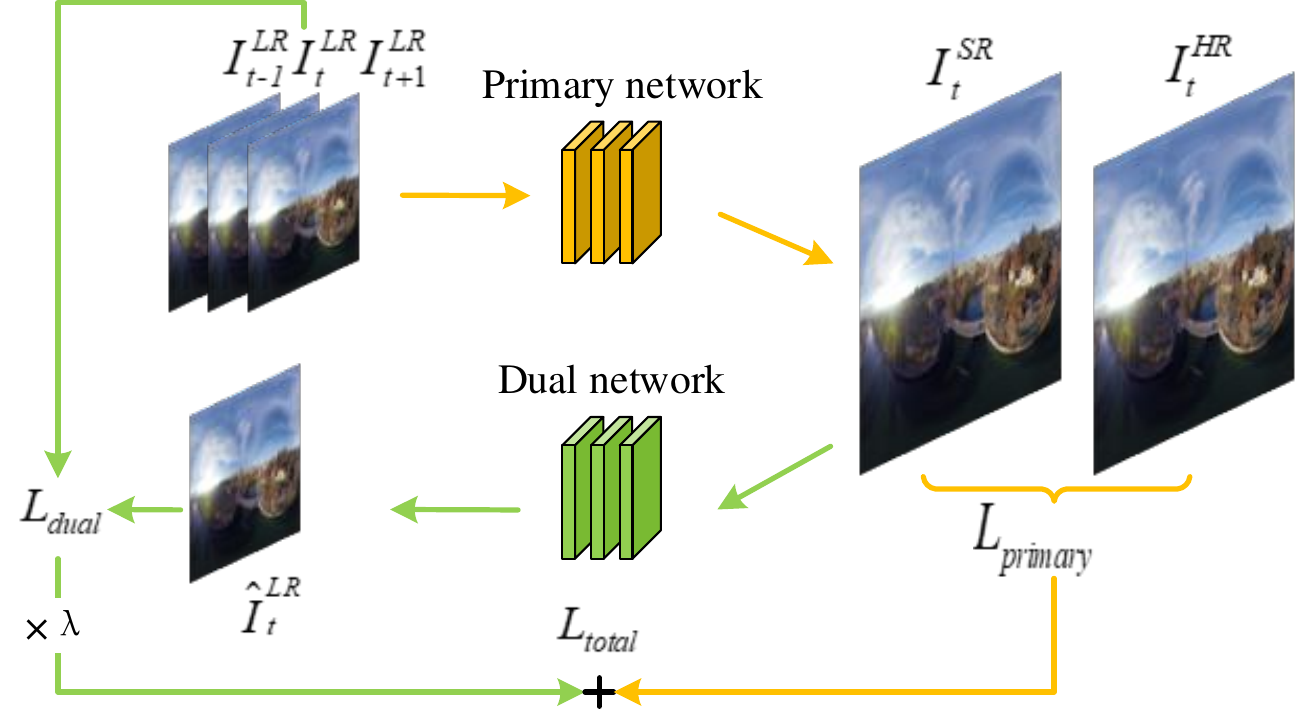}
\caption{Learning process of the proposed dual mechanism. Note: the primary network is our main super-resolution network (including single frame network and multi-frame network).}
\label{dual_lg}
\end{figure}


\subsection{Dual Network}
During the training stage, a dual network is devised to constrain the solution space, as shown in Fig.\ \ref{dual_lg}. First of all, the dual network converts the SR image into the LR space. We denote the output of the dual network as $\hat I_t^{LR}$.
An extra loss function $L_{dual}$ is defined by computing the weighted mean squared error between the original LR image $I_t^{LR}$ and $\hat I_t^{LR}$.
By using this loss function, the solution space can be constrained effectively, and thus a better solution might be found. In our experiments, we adopts two 3$\times$3 convolutional layers with stride 2 as our dual network.

\begin{figure*}[ht]
\centering
\includegraphics[scale=0.7]{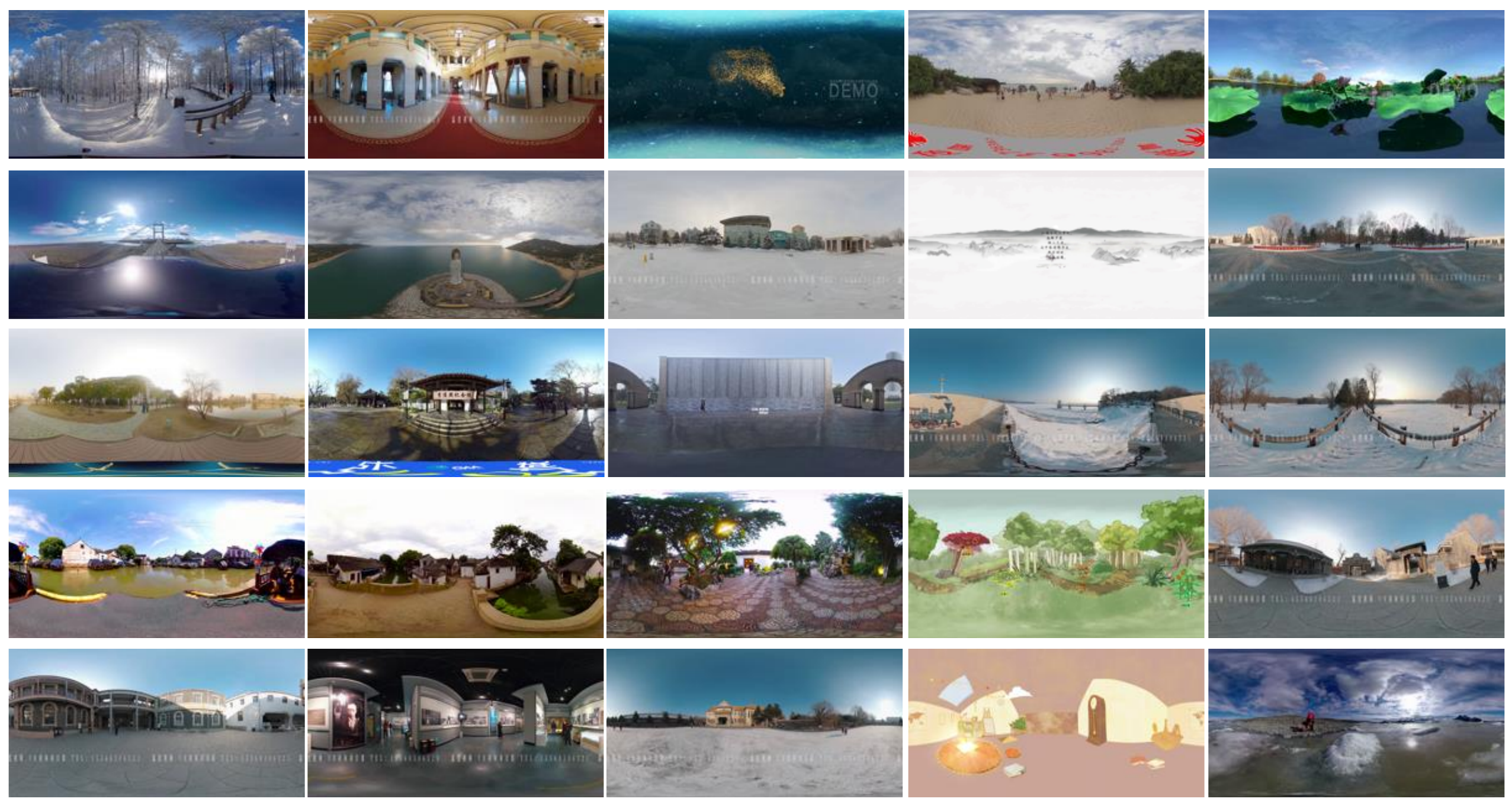}
\caption{Examples of our MiG panorama video dataset.}
\label{dataset}
\end{figure*}

\subsection{Loss Function}
\label{sec4.5}
In virtual reality, we generally experience the \ang{360} panorama videos by a special VR display device such as a head-mounted one. For the panorama video in ERP format, its important contents are generally displayed in the equatorial regions. Therefore, we should pay more attention to process these regions. In our method, we achieve this goal by assigning different weight values for different pixels according to their latitudes, and pixels in lower latitudes can get larger weights. Based on the above analysis, we utilize a new loss function, named weighted mean square error (WMSE). Then the loss containing the single frame network and the multi-frame network $L_{primary}$ is defined as follows:
\begin{equation}
\label{wmse}
{\frac{1}{\sum_{i=0}^{M-1}\!\sum_{j=0}^{N-1}\!w(i,j)}\sum_{i=0}^{M-1}\sum_{j=0}^{N-1}w(i.j){\cdot}(I_t^{\textup{SR}}(i,j)-}{I_t^{\textup{HR}}(i,j))^2}
\end{equation}
where $M$ and $N$ denote the width and height of an image, respectively. Here, $(i,j)$ denotes the pixel coordinate, and $w(i,j)$ is the weight value at this location defined as
\begin{equation}
w(i,j) = \cos\frac{(j+0.5-N/2)\pi}{N}.
\end{equation}
Then the total training loss $L_{total}$ of the entire network is defined as follows:
\begin{equation}
L_{total} = L_{primary} + \lambda L_{dual}
\end{equation}
where $L_{dual}$ is computed as in Eq.\ (\ref{wmse}), and it denotes the loss function constraining the output of the dual network. Here  $\lambda$ is a balance factor between $L_{primary}$ and $L_{dual}$.

Finally, we compare the network architecture of  SMFN with those of the state-of-the-art video super-resolution methods in terms of network architectures, as shown in Table \ref{arch_cmp}, and highlight the novelty of our SMFN method. In other words, our SMFN method is the first work for super-resolution of \ang{360} panorama videos.



\begin{table}[h]
\caption{Comparison of network architectures used in the state-of-the-art methods and our method.}
\label{arch_cmp}
\centering
\begin{tabular}{cccc}
\hline
Methods&\!\!\!\!Single frame modual\!\!&\!\!Multi-frame modual\!\!&\!\!Dual training\!\!\\
		\hline
		EDVR \cite{EDVR}&$\times$&\checkmark&$\times$\\
		RBPN \cite{RBPN}&\checkmark&\checkmark&$\times$\\
		\!\!SOFVSR \cite{SOFVSR,SOFVSR-TIP}\!\!&$\times$&\checkmark&$\times$\\
		FRVSR \cite{FRVSR}&$\times$&\checkmark&$\times$\\
		VESPCN \cite{VESPCN}&$\times$&\checkmark&$\times$\\
		VSRnet \cite{VSRnet}&$\times$&\checkmark&$\times$\\	
		SMFN (ours)&\checkmark&\checkmark&\checkmark\\
		\hline
	\end{tabular}
\end{table}

\begin{table}[h]
\caption{Ablation studies of different modules in our network SMFN.}
\label{ablation}
\centering
\begin{tabular}{cccc}
\hline
&WS-PSNR&WS-SSIM&Params.(MB)\\
\hline
Baseline (SMFN)&\textbf{30.13}&\textbf{0.8381}&16.36\\
SMFN w/o attention&30.10&0.8379&15.93\\
		SMFN w/o alignment &30.08&0.8375&14.98\\
		SMFN w/o dual network&30.10&0.8367&16.36\\
        SMFN w/o fusion module&30.07&0.8361&16.36\\
		SMFN w/o single frame network&29.96&0.8344&5.95\\			
		\hline
	\end{tabular}
\end{table}


\section{Experimental Results}
In this section, we propose the MiG Panoramic Video dataset for super-resolution of panorama videos. Then we conduct ablation studies, and evaluate the performance our SMFN on this dataset.

\subsection{Datasets}
To the best of our knowledge, there are no publicly available datasets for the super-resolution of panorama videos. Therefore, we collect and edit a dataset named MiG Panorama Video from the Internet. It includes 204 panorama videos in total in the Equirectangular projection (ERP) format which is one of the most widely used projection scheme for panoramic content, and each video includes 100 frames with resolution ranging from 4,096$\times$2,048 to 1,440$\times$720. The scenes vary from outdoor to indoor, and from day to night, as shown in Fig.\ \ref{dataset}. Then we select 4 representative videos as test set and the others are used as training set. Then considering the limitation of computing resources, we downsample each video with scale factor 2 to resolution ranging from 2,048$\times$1024 to 720$\times$360 using the bicubic interpolation algorithm as the ground truth. Then we further downsample the ground truth videos with scale factor 4 using the same interpolation algorithm to obtain the corresponding LR videos. Our dataset are released publicly for researchers to further study super-resolution of panorama videos.

As suggested in \cite{ye2017algorithm}, the Weighted to Spherically uniform-Peak Signal To Noise Ratio (WS-PSNR) and WS-Structural Similarity Index Measure (WS-SSIM) are used as metrics to evaluate the performance of all the algorithms for panorama videos super-resolution. Since most of other compared algorithms are for the super-resolution of generic videos, we also report the PSNR and SSIM results of all the algorithms.

\begin{table*}[t]
\centering
\caption{Quantitative comparison of the representative algorithms and our SMFN method in terms of both WS-PSNR (top) and WS-SSIM (bottom). Note that the best performance is shown in red, and the second best performance is shown in blue.}
\label{cmp}
\begin{tabular}{ccccccccccc}
\hline
{Clips}&{\!Bicubic\!}&{\!\!SR360 \cite{sr360}\!\!}&{\!\!\!VSRnet \cite{VSRnet}\!\!\!}&{\!\!\!VESPCN \cite{VESPCN}\!\!\!}&{\!\!\!FRVSR \cite{FRVSR}\!\!\!}&{\!\!\!SOFVSR \cite{SOFVSR}\!\!\!}&{\!\!\!SOFVSR-TIP \cite{SOFVSR-TIP}\!\!\!}&{\!\!\!RBPN \cite{RBPN}\!\!\!}&{\!\!\!EDVR \cite{EDVR}\!\!\!}&{\!\!SMFN (ours)\!\!}\\
\hline
\multirow{2}*{Clip\_1}&	26.39&26.62&	26.60&26.75&26.74&26.76&26.77&26.76&\color{blue}{26.80}&\color{red}{26.84}\\
		&0.6888&0.7131&0.7118&0.7257&0.7263&0.7264&	0.7257&0.7270&\color{blue}{0.7293}&\color{red}{0.7305}\\
\hline
\multirow{2}*{Clip\_2}&	28.64&29.37&	28.94&29.63&29.58&29.66&29.74&29.86&\color{blue}{30.04}&\color{red}{30.20}\\
		&0.8274&0.8422&0.8386&0.8574&0.8569&0.8587&	0.8622&0.8671&\color{blue}{0.8744}&\color{red}{0.8779}\\
		\hline
\multirow{2}*{Clip\_3}&	29.75&30.76&	30.15&31.24&31.20&31.28&31.29&31.51&\color{blue}{31.57}&\color{red}{31.92}\\
		&0.8009&0.8214&0.8165&0.8374&0.8379&0.8378&	0.8392&0.8432&\color{blue}{0.8464}&\color{red}{0.8493}\\
		\hline
\multirow{2}*{Clip\_4}&	30.46&30.85&	30.72&31.19&30.99&31.21&31.24&31.30&\color{blue}{31.43}&\color{red}{31.57}\\
		&0.8685&0.8726&0.8779&0.8869&0.8854&0.8869&	0.8880&0.8904&\color{blue}{0.8929}&\color{red}{0.8948}\\
		\hline
\multirow{2}*{Average}&	28.81&29.40&	29.10&29.70&29.63&29.73&29.76&29.86&\color{blue}{29.96}&\color{red}{30.13}\\
		&0.7964&0.8123&0.8112&0.8268&0.8266&0.8274&	0.8288&0.8319&\color{blue}{0.8358}&\color{red}{0.8381}\\
\hline
\end{tabular}
\end{table*}

\begin{table*}[t]
\centering
\caption{Quantitative comparison of the representative algorithms and our SMFN method in terms of PSNR (top) and SSIM (bottom). Note that the best performance is shown in red, and the second best performance is shown in blue.}
\label{cmp1}	
\begin{tabular}{ccccccccccc}
\hline		
{Clips}&{\!Bicubic\!}&{\!\!SR360 \cite{sr360}\!\!}&{\!\!\!VSRnet \cite{VSRnet}\!\!\!}&{\!\!\!VESPCN \cite{VESPCN}\!\!\!}&{\!\!\!FRVSR \cite{FRVSR}\!\!\!}&{\!\!\!SOFVSR \cite{SOFVSR}\!\!\!}&{\!\!\!SOFVSR-TIP \cite{SOFVSR-TIP}\!\!\!}&{\!\!\!RBPN \cite{RBPN}\!\!\!}&{\!\!\!EDVR \cite{EDVR}\!\!\!}&{\!\!SMFN (ours)\!\!}\\
\hline
\multirow{2}*{Clip\_1}&	26.38&26.58&	26.59&26.71&26.70&26.72&26.72&26.71&\color{blue}{26.73}&\color{red}{26.78}\\
		&0.6868&0.7101&0.7075&0.7203&0.7215&0.7213&	0.7203&\color{blue}{0.7218}&0.7217&\color{red}{0.7251}\\
\hline
\multirow{2}*{Clip\_2}&	30.09&30.69&	30.39&31.04&30.99&31.08&31.16&31.26&\color{blue}{31.48}&\color{red}{31.60}\\
		&0.8494&0.8580&0.8573&0.8723&0.8718&0.8744&	0.8775&0.8810&\color{blue}{0.8883}&\color{red}{0.8904}\\
\hline
\multirow{2}*{Clip\_3}&	27.65&29.29&	28.10&29.50&29.43&29.53&29.54&29.92&\color{blue}{30.18}&\color{red}{30.82}\\
		&0.8119&0.8406&0.8245&0.8490&0.8502&0.8500&	0.8527&0.8589&\color{blue}{0.8630}&\color{red}{0.8660}\\
\hline
\multirow{2}*{Clip\_4}&	31.88&32.22&	32.15&32.63&32.41&32.67&32.70&32.76&\color{blue}{32.91}&\color{red}{33.05}\\
		&0.9005&0.9001&0.9069&0.9134&0.9121&0.9138&	0.9147&0.9169&\color{blue}{0.9186}&\color{red}{0.9204}\\
\hline
\multirow{2}*{Average\!}&	29.00&29.69&	29.30&29.97&29.88&30.00&30.03&30.16&\color{blue}{30.32}&\color{red}{30.56}\\
		&0.8121&0.8272&0.8241&0.8388&0.8389&0.8399&	0.8413&0.8446&\color{blue}{0.8479}&\color{red}{0.8505}\\
\hline
\end{tabular}
\end{table*}

\subsection{Training Setting}
In our single frame network, we use 32 layers of convolutions. The feature extraction module adopts 3 residual blocks. In the reconstruction module, we employ 5 residual dense blocks. The channel size in each residual block is set to 64. The channel size in each residual dense block is set to 64 with a growth rate of 32.  In our experiments, we first convert the RGB videos to the YCbCr space and then use the Y channel as the input of our network. The network takes three consecutive frames (i.e., $N\!=\!1$) as input unless otherwise specified. During the training stage, we use patches with size of $32\times32$ as input, and the batch size is set to 16. In addition, we use mainly geometric augmentation techniques including reflection, random cropping, and rotation.


For the primary network, we adopts the Adam optimizer with $\beta_1=0.9$ and $\beta_2=0.999$ to train the network. The initial learning rate is set to $1\times10^{-4}$. Then we decay the learning rate to half after every 20 epoches. The configuration of the dual network is the same as that of the single and multi-frame network. The value of the parameter $\lambda$ in our loss function is set to 0.1. We implement our models in PyTorch framework and train them using 2 NVIDIA Titan Xp GPUs.

\subsection{Ablation Study}
In this subsection, we analyze the contributions of each module in our network, including a dual network, an attention module, a single frame network, a fusion module, and an alignment module, as listed in Table \ref{ablation}. The proposed network is used as a baseline. The performance (including WS-PSNR and WS-SSIM) of our method without the attention module drops by 0.03dB and 0.0002, respectively, compared with our baseline. Without the alignment module, the WS-PSNR and WS-SSIM results fall by 0.05 dB and 0.0006, respectively. The ablation of the dual network is similar to that of the alignment module. Our network without the fusion module drops by 0.06 dB and 0.0018, respectively. If our network does not include the single frame network, the WS-PSNR and WS-SSIM results drop by 0.17 dB and 0.0037 compared with the baseline. This indicates that the single frame network plays an important role in the enhancement of the WS-PSNR but at a cost of 10.41MB parameters. The main cost may probably be dominated by the 32 layers of convolution. Moreover, the fusion module gains relative higher WS-PSNR while without increasing parameters. The dual network also enhances the performance but without raising the number of parameters.


\subsection{Quantitative Comparison}
In this subsection, we compare our SMFN method with the state-of-the-art super-resolution methods in terms of WS-PSNR and WS-SSIM on the Y channel. The compared methods include SR360 \cite{sr360}, VSRnet \cite{VSRnet}, VESPCN \cite{VESPCN}, FRVSR \cite{FRVSR}, SOFVSR \cite{SOFVSR,SOFVSR-TIP}, RBPN \cite{RBPN} and EDVR \cite{EDVR}. For fair comparison, we re-train the methods from scratch except for SR360. Because this is a single frame panorama image super-resolution method, we use its pre-trained model for testing. The quantitative results (including WS-PSNR and WS-SSIM) of all the methods are reported in Table \ref{cmp}. All the results show that the best performance is obtained by the proposed method, which demonstrates the effectiveness of our SMFN.

Moreover, we compare our SMFN method with state-of-the-art methods in terms of PSNR and SSIM, which are the widely used evaluation criterion for super-resolution. The experimental results are shown in Table \ref{cmp1}. All the results indicate that the results measured by PSNR and SSIM are consistent with those of WS-PSNR and WS-SSIM, respectively. Moreover, our SMFN method outperforms the second best method EDVR with large margins (0.17-0.24dB) in terms of WS-PSNR and PSNR. The experimental results indicate that our SMFN method has a better modeling ability for \ang{360} panorama videos compared with other video super-resolution methods.

\subsection{Qualitative Comparison}
\begin{figure*}
 \centering
\includegraphics[scale=0.6]{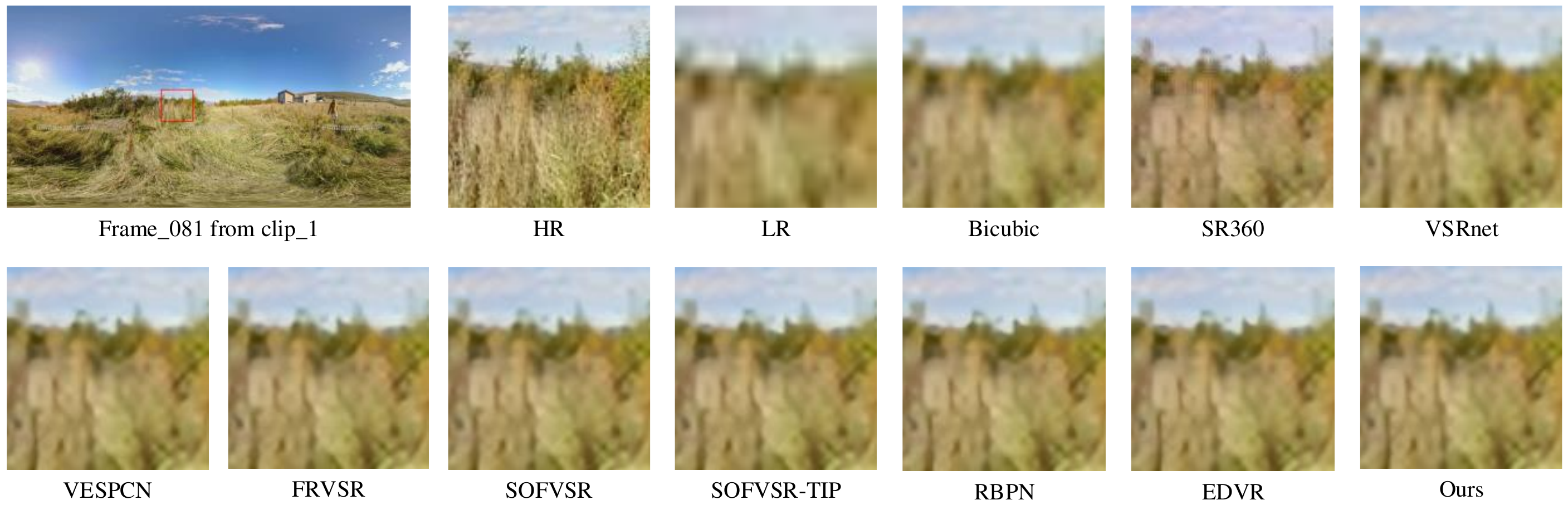}
\includegraphics[scale=0.6]{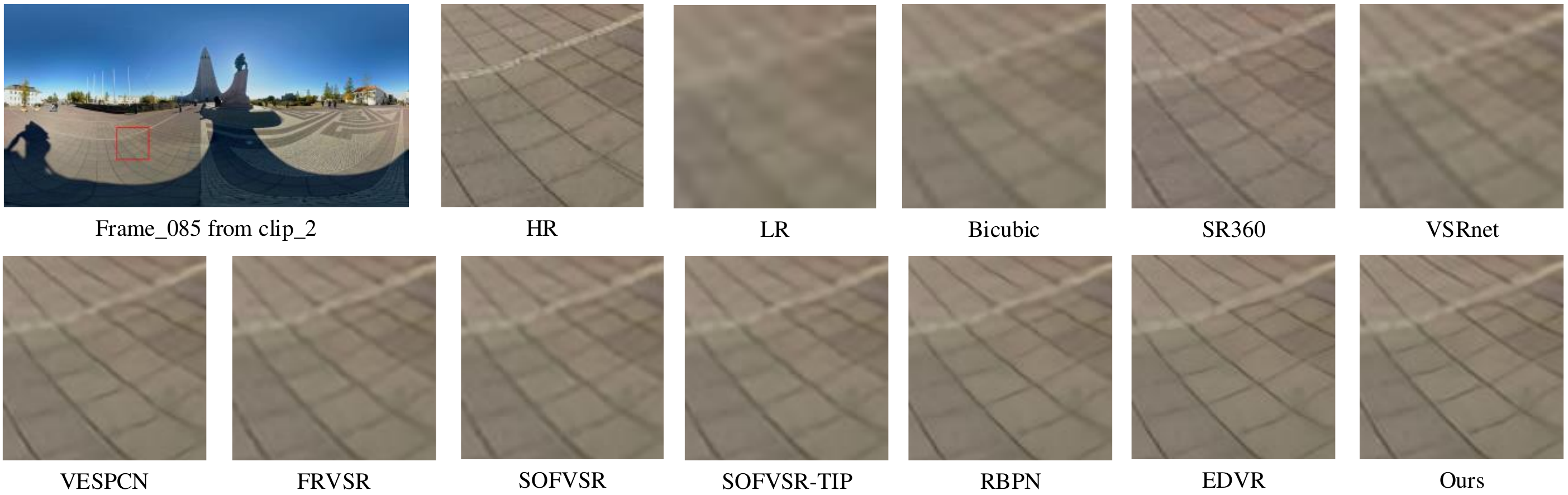}
\includegraphics[scale=0.6]{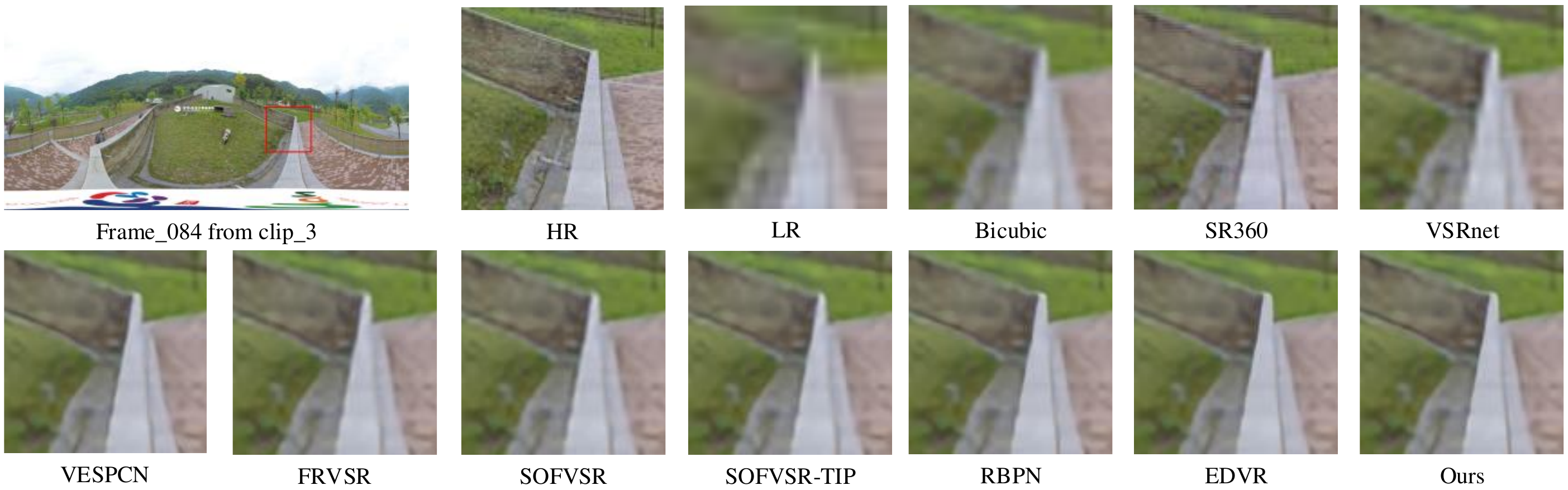}
\includegraphics[scale=0.6]{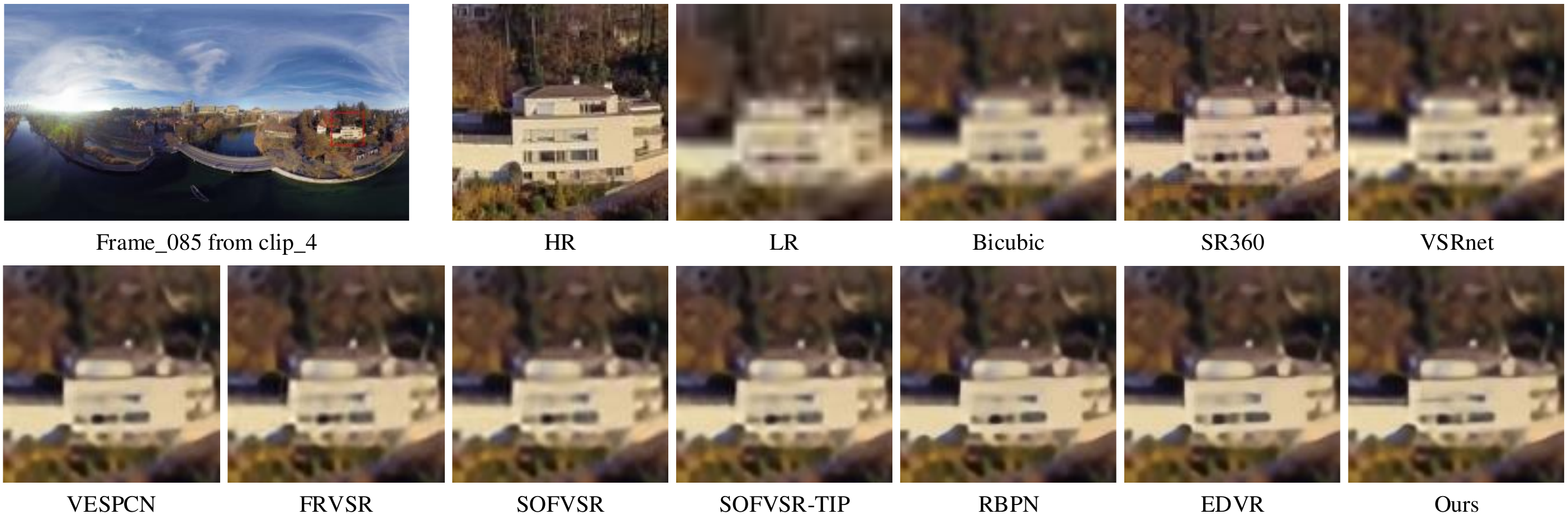}
\caption{Visual comparisons of different methods on four images from the MiG panorama video dataset for $\times4$ scale. The methods include Bicubic, SR360 \cite{sr360}, VSRnet \cite{VSRnet}, VESPCN \cite{VESPCN}, FRVSR \cite{FRVSR}, SOFVSR \cite{SOFVSR}, SOFVSR-TIP \cite{SOFVSR-TIP}, RBPN \cite{RBPN}, EDVR \cite{EDVR}, and SMFN (ours).}
\label{compare}
\end{figure*}

In this subsection, we show the qualitative comparison of the state-of-the-art methods and our SMFN method, as shown in Fig.\ \ref{compare}. For the image frame\_084 from clip\_3, our method recovers more texture details on the wall, while the results of other algorithms are blurring. For the image frame\_085 from clip\_4, SR360 \cite{sr360} yields a HR image distorted in color, while our method is able to maintain superior consistency. This phenomenon indicates that super-resolution algorithms based on GAN generally produce some nonexistent details, which might impair the visual performance. We can also see that in the image frame\_085 from clip\_2 and the image frame\_081 from clip\_1, SR360 \cite{sr360} produces artifacts of different degrees that severely affect our visual experience. On the contrary, our SMFN method can recover better results in terms of both quantitative metrics and qualitative results. Overall, compared with other methods, our SMFN method generates fewer artifacts and richer high-frequency details in the super-resolution of  \ang{360} panorama videos.

\section{Conclusions and Future Work}
In this paper, we proposed a novel single frame and multi-frame joint network for super-resolution of  \ang{360} panorama videos. The proposed method uses deformable convolution to align the target features with the neighboring features, mixed mechanism to enhance the feature representation capability, and dual learning mechanism to constrain the space of solution so that a better solution can be found. A novel loss function based on weighted mean square error (WMSE) was also proposed to emphasize on the recovering quality of the equatorial regions, which usually have more important visual content in \ang{360} panorama videos. Compared with state-of-the-art super-resolution algorithms developed for standard videos, the proposed method performs much better in terms of both quantitative and qualitative results.

In the future, we will investigate fast and lightweight video super-resolution methods as in \cite{fastnet} for panorama videos, which can be applied in real-word applications, such as lightweight \ang{360} panorama video super-resolution algorithms. In this work, we mainly focus on the case of the magnification factor $4$, and thus we will study some more challenging scales such as $\times8$ and $\times16$ in the future. In addition, video super-resolution for higher resolution panorama videos (e.g., 8K and 16K) is also one future work.

\section*{Acknowledgments}
This work was supported by the National Natural Science Foundation of China (Nos.\ 61876220, 61876221, 61976164, 61836009 and U1701267, and~61871310), the Project supported the Foundation for Innovative Research Groups of the National Natural Science Foundation of China (No.\ 61621005), the Program for Cheung Kong Scholars and Innovative Research Team in University (No.\ IRT\_15R53), the Fund for Foreign Scholars in University Research and Teaching Programs (the 111 Project) (No.\ B07048), the Science Foundation of Xidian University (Nos.\ 10251180018 and 10251180019), the National Science Basic Research Plan in Shaanxi Province of China (Nos.\ 2019JQ-657 and 2020JM-194), and the Key Special Project of China High Resolution Earth Observation System-Young Scholar Innovation Fund.

\ifCLASSOPTIONcaptionsoff
  \newpage
\fi




%
\bibliographystyle{IEEEtran}
\bibliography{IEEEabrv,reference}

\begin{thebibliography}{10}
\providecommand{\url}[1]{#1}
\csname url@samestyle\endcsname
\providecommand{\newblock}{\relax}
\providecommand{\bibinfo}[2]{#2}
\providecommand{\BIBentrySTDinterwordspacing}{\spaceskip=0pt\relax}
\providecommand{\BIBentryALTinterwordstretchfactor}{4}
\providecommand{\BIBentryALTinterwordspacing}{\spaceskip=\fontdimen2\font plus
\BIBentryALTinterwordstretchfactor\fontdimen3\font minus
  \fontdimen4\font\relax}
\providecommand{\BIBforeignlanguage}[2]{{%
\expandafter\ifx\csname l@#1\endcsname\relax
\typeout{** WARNING: IEEEtran.bst: No hyphenation pattern has been}%
\typeout{** loaded for the language `#1'. Using the pattern for}%
\typeout{** the default language instead.}%
\else
\language=\csname l@#1\endcsname
\fi
#2}}
\providecommand{\BIBdecl}{\relax}
\BIBdecl

\bibitem{panorama1}
A.~{Rana}, C.~{Ozcinar}, and A.~{Smolic}, ``Towards generating ambisonics using
  audio-visual cue for virtual reality,'' in \emph{IEEE Int. Conf. Acoust.
  Speech Signal Process. Proc. (ICASSP)}, 2019, pp. 2012--2016.

\bibitem{panorama2}
C.~{Ozcinar}, J.~{Cabrera}, and A.~{Smolic}, ``Visual attention-aware
  omnidirectional video streaming using optimal tiles for virtual reality,''
  \emph{IEEE J. Emerg. Sel. Top. Circuits Syst.}, vol.~9, no.~1, pp. 217--230,
  Mar. 2019.

\bibitem{panorama3}
C.~{Ozcinar}, A.~{De Abreu}, and A.~{Smolic}, ``Viewport-aware adaptive
  360$^\circ$ video streaming using tiles for virtual reality,'' in \emph{Proc.
  Int. Conf. Image Process. (ICIP)}, 2017, pp. 2174--2178.

\bibitem{panorama5}
M.~S. {Elbamby}, C.~{Perfecto}, M.~{Bennis}, and K.~{Doppler}, ``Toward
  low-latency and ultra-reliable virtual reality,'' \emph{IEEE Netw.}, vol.~32,
  no.~2, pp. 78--84, Apr. 2018.

\bibitem{RBPN}
M.~{Haris}, G.~{Shakhnarovich}, and N.~{Ukita}, ``Recurrent back-projection
  network for video super-resolution,'' in \emph{Proc. IEEE Conf. Comput. Vis.
  Pattern Recognit. (CVPR)}, 2019, pp. 3892--3901.

\bibitem{EDVR}
X.~{Wang}, K.~C.~K. {Chan}, K.~{Yu}, C.~{Dong}, and C.~C. {Loy}, ``Edvr: Video
  restoration with enhanced deformable convolutional networks,'' in \emph{IEEE
  Comput. Soc. Conf. Comput. Vis. Pattern Recogn. Workshops}, 2019, pp.
  1954--1963.

\bibitem{wang2020deep}
Z.~Wang, J.~Chen, and S.~C. Hoi, ``Deep learning for image super-resolution: A
  survey,'' \emph{IEEE Trans. Pattern Anal. Mach. Intell.}, 2020.

\bibitem{SRCNN}
C.~Dong, C.~C. Loy, K.~He, and X.~Tang, ``Learning a deep convolutional network
  for image super-resolution,'' in \emph{Lect. Notes Comput. Sci.}, D.~Fleet,
  T.~Pajdla, B.~Schiele, and T.~Tuytelaars, Eds.\hskip 1em plus 0.5em minus
  0.4em\relax Cham: Springer International Publishing, 2014, pp. 184--199.

\bibitem{FSRCNN}
C.~Dong, C.~C. Loy, and X.~Tang, ``Accelerating the super-resolution
  convolutional neural network,'' in \emph{Lect. Notes Comput. Sci.}, B.~Leibe,
  J.~Matas, N.~Sebe, and M.~Welling, Eds.\hskip 1em plus 0.5em minus
  0.4em\relax Cham: Springer International Publishing, 2016, pp. 391--407.

\bibitem{ESPCN}
W.~{Shi}, J.~{Caballero}, F.~{Huszár}, J.~{Totz}, A.~P. {Aitken}, R.~{Bishop},
  D.~{Rueckert}, and Z.~{Wang}, ``Real-time single image and video
  super-resolution using an efficient sub-pixel convolutional neural network,''
  in \emph{Proc. IEEE Conf. Comput. Vis. Pattern Recognit (CVPR)}, 2016, pp.
  1874--1883.

\bibitem{EDSR}
B.~{Lim}, S.~{Son}, H.~{Kim}, S.~{Nah}, and K.~M. {Lee}, ``Enhanced deep
  residual networks for single image super-resolution,'' in \emph{IEEE Comput.
  Soc. Conf. Comput. Vis. Pattern Recogn. Workshops}, 2017, pp. 1132--1140.

\bibitem{RCAN}
Y.~Zhang, K.~Li, K.~Li, L.~Wang, B.~Zhong, and Y.~Fu, ``Image super-resolution
  using very deep residual channel attention networks,'' in \emph{Lect. Notes
  Comput. Sci.}, 2018, pp. 286--301.

\bibitem{ResNet}
K.~{He}, X.~{Zhang}, S.~{Ren}, and J.~{Sun}, ``Deep residual learning for image
  recognition,'' in \emph{Proc. IEEE Conf. Comput. Vis. Pattern Recognit.
  (CVPR)}, 2016, pp. 770--778.

\bibitem{autoencoder}
K.~Zeng, J.~Yu, R.~Wang, C.~Li, and D.~Tao, ``Coupled deep autoencoder for
  single image super-resolution,'' \emph{IEEE Trans. Cybernetics}, vol.~47,
  no.~1, pp. 27--37, Jan. 2020.

\bibitem{DBPN}
M.~{Haris}, G.~{Shakhnarovich}, and N.~{Ukita}, ``Deep back-projection networks
  for super-resolution,'' in \emph{Proc. IEEE Conf. Comput. Vis. Pattern
  Recognit. (CVPR)}, 2018, pp. 1664--1673.

\bibitem{RDN}
Y.~{Zhang}, Y.~{Tian}, Y.~{Kong}, B.~{Zhong}, and Y.~{Fu}, ``Residual dense
  network for image super-resolution,'' in \emph{Proc. IEEE Conf. Comput. Vis.
  Pattern Recognit. (CVPR)}, 2018, pp. 2472--2481.

\bibitem{NLN}
X.~{Wang}, R.~{Girshick}, A.~{Gupta}, and K.~{He}, ``Non-local neural
  networks,'' in \emph{Proc. IEEE Conf. Comput. Vis. Pattern Recognit. (CVPR)},
  2018, pp. 7794--7803.

\bibitem{res_nonlocal}
Y.~Zhang, K.~Li, K.~Li, B.~Zhong, and Y.~Fu, ``Residual non-local attention
  networks for image restoration,'' in \emph{Int. Conf. Learn. Represent.
  (ICLR)}, 2019.

\bibitem{nonlocal_rcnn}
D.~Liu, B.~Wen, Y.~Fan, C.~C. Loy, and T.~S. Huang, ``Non-local recurrent
  network for image restoration,'' in \emph{Adv. neural inf. proces. syst.},
  2018, pp. 1673--1682.

\bibitem{CBAM}
S.~Woo, J.~Park, J.-Y. Lee, and I.~So~Kweon, ``Cbam: Convolutional block
  attention module,'' in \emph{Lect. Notes Comput. Sci.}, September 2018.

\bibitem{SAN}
T.~Dai, J.~Cai, Y.~Zhang, S.-T. Xia, and L.~Zhang, ``Second-order attention
  network for single image super-resolution,'' in \emph{Proc. IEEE Conf.
  Comput. Vis. Pattern Recognit (CVPR)}, 2019.

\bibitem{FB}
A.~R. Zamir, T.-L. Wu, L.~Sun, W.~B. Shen, B.~E. Shi, J.~Malik, and
  S.~Savarese, ``Feedback networks,'' in \emph{Proc. IEEE Conf. Comput. Vis.
  Pattern Recognit (CVPR)}, 2017.

\bibitem{SRFBN}
Z.~Li, J.~Yang, Z.~Liu, X.~Yang, G.~Jeon, and W.~Wu, ``Feedback network for
  image super-resolution,'' in \emph{Proc. IEEE Conf. Comput. Vis. Pattern
  Recognit (CVPR)}, 2019.

\bibitem{dual_1}
D.~He, Y.~Xia, T.~Qin, L.~Wang, N.~Yu, T.-Y. Liu, and W.-Y. Ma, ``Dual learning
  for machine translation,'' in \emph{Adv. neural inf. proces. syst.}, 2016,
  pp. 820--828.

\bibitem{dual_2}
Y.~Xia, X.~Tan, F.~Tian, T.~Qin, N.~Yu, and T.-Y. Liu, ``Model-level dual
  learning,'' in \emph{Int. Conf. Mach. Learn. (ICML)}, 2018, pp. 5383--5392.

\bibitem{dual_3}
Y.~Zhang, T.~Xiang, T.~M. Hospedales, and H.~Lu, ``Deep mutual learning,'' in
  \emph{Proc. IEEE Conf. Comput. Vis. Pattern Recognit (CVPR)}, 2018.

\bibitem{dual_4}
Y.~Xia, T.~Qin, W.~Chen, J.~Bian, N.~Yu, and T.-Y. Liu, ``Dual supervised
  learning,'' in \emph{Int. Conf. Mach. Learn. (ICML)}, 2017, pp. 3789--3798.

\bibitem{DRN}
Y.~Guo, J.~Chen, J.~Wang, Q.~Chen, J.~Cao, Z.~Deng, Y.~Xu, and M.~Tan,
  ``Closed-loop matters: Dual regression networks for single image
  super-resolution,'' in \emph{Proc. IEEE Conf. Comput. Vis. Pattern Recognit
  (CVPR)}, 2020.

\bibitem{sr360}
C.~{Ozcinar}, A.~{Rana}, and A.~{Smolic}, ``Super-resolution of omnidirectional
  images using adversarial learning,'' in \emph{IEEE Int. Workshop Multimed.
  Signal Process. (MMSP)}, Sep 2019.

\bibitem{vrsrcnn2018}
V.~Fakour-Sevom, E.~Guldogan, and J.-K. K{\"a}m{\"a}r{\"a}inen, ``360 panorama
  super-resolution using deep convolutional networks,'' in \emph{VISIGRAPP -
  Proc. Int. Jt. Conf. Comput. Vis., Imaging Comput. Graph. Theory Appl.},
  vol.~1, 2018.

\bibitem{vrw2020}
S.~{Li}, C.~{Lin}, K.~{Liao}, Y.~{Zhao}, and X.~{Zhang}, ``Panoramic image
  quality-enhancement by fusing neural textures of the adaptive initial
  viewport,'' in \emph{Proc. IEEE Conf. Virtual Real. 3D User Interfaces, VRW},
  2020, pp. 816--817.

\bibitem{OurSurvey}
H.~Liu, Z.~Ruan, P.~Zhao, F.~Shang, L.~Yang, and Y.~Liu, ``Video super
  resolution based on deep learning: A comprehensive survey,'' \emph{arXiv
  preprint arXiv:2007.12928}, 2020.

\bibitem{VSRnet}
A.~Kappeler, S.~Yoo, Q.~Dai, and A.~K. Katsaggelos, ``Video super-resolution
  with convolutional neural networks,'' \emph{IEEE Trans. Comput. Imaging},
  vol.~2, no.~2, pp. 109--122, 2016.

\bibitem{VESPCN}
J.~{Caballero}, C.~{Ledig}, A.~{Aitken}, A.~{Acosta}, J.~{Totz}, Z.~{Wang}, and
  W.~{Shi}, ``Real-time video super-resolution with spatio-temporal networks
  and motion compensation,'' in \emph{Proc. IEEE Conf. Comput. Vis. Pattern
  Recognit. (CVPR)}, 2017, pp. 2848--2857.

\bibitem{TOFlow}
T.~Xue, B.~Chen, J.~Wu, D.~Wei, and W.~T. Freeman, ``Video enhancement with
  task-oriented flow,'' \emph{Int. J. Comput. Vis.}, vol. 127, no.~8, pp.
  1106--1125, 2019.

\bibitem{FRVSR}
M.~S.~M. {Sajjadi}, R.~{Vemulapalli}, and M.~{Brown}, ``Frame-recurrent video
  super-resolution,'' in \emph{Proc. IEEE Conf. Comput. Vis. Pattern Recognit.
  (CVPR)}, 2018, pp. 6626--6634.

\bibitem{VSRResNet}
A.~Lucas, S.~Lopez-Tapia, R.~Molina, and A.~K. Katsaggelos, ``Generative
  adversarial networks and perceptual losses for video super-resolution,''
  \emph{IEEE Trans. Image Process.}, vol.~28, no.~7, pp. 3312--3327, 2019.

\bibitem{DUF}
Y.~{Jo}, S.~W. {Oh}, J.~{Kang}, and S.~J. {Kim}, ``Deep video super-resolution
  network using dynamic upsampling filters without explicit motion
  compensation,'' in \emph{Proc. IEEE Conf. Comput. Vis. Pattern Recognit.
  (CVPR)}, 2018, pp. 3224--3232.

\bibitem{FSTRN}
S.~Li, F.~He, B.~Du, L.~Zhang, Y.~Xu, and D.~Tao, ``Fast spatio-temporal
  residual network for video super-resolution,'' in \emph{Proc. IEEE Conf.
  Comput. Vis. Pattern Recognit. (CVPR)}, 2019, pp. 10\,522--10\,531.

\bibitem{FFCVSR}
B.~Yan, C.~Lin, and W.~Tan, ``Frame and feature-context video
  super-resolution,'' in \emph{Proc. Conf. Artif. Intell., AAAI}.\hskip 1em
  plus 0.5em minus 0.4em\relax {AAAI} Press, 2019, pp. 5597--5604.

\bibitem{3DSRnet}
S.~Y. {Kim}, J.~{Lim}, T.~{Na}, and M.~{Kim}, ``Video super-resolution based on
  3d-cnns with consideration of scene change,'' in \emph{Proc. Int. Conf. Image
  Process. (ICIP)}, 2019, pp. 2831--2835.

\bibitem{DRVSR}
X.~{Tao}, H.~{Gao}, R.~{Liao}, J.~{Wang}, and J.~{Jia}, ``Detail-revealing deep
  video super-resolution,'' in \emph{Proc. IEEE Int. Conf. Comput. Vision
  (ICCV)}, 2017, pp. 4482--4490.

\bibitem{D3Dnet}
X.~Ying, L.~Wang, Y.~Wang, W.~Sheng, W.~An, and Y.~Guo, ``Deformable 3d
  convolution for video super-resolution,'' \emph{IEEE Signal Processing
  Letters}, 2020.

\bibitem{DCN}
J.~{Dai}, H.~{Qi}, Y.~{Xiong}, Y.~{Li}, G.~{Zhang}, H.~{Hu}, and Y.~{Wei},
  ``Deformable convolutional networks,'' in \emph{Proc. IEEE Int. Conf. Comput.
  Vision (ICCV)}, 2017, pp. 764--773.

\bibitem{DCNV2}
X.~{Zhu}, H.~{Hu}, S.~{Lin}, and J.~{Dai}, ``Deformable convnets v2: More
  deformable, better results,'' in \emph{Proc. IEEE Conf. Comput. Vis. Pattern
  Recognit. (CVPR)}, 2019, pp. 9300--9308.

\bibitem{resnet2}
R.~Lan, L.~Sun, Z.~Liu, H.~Lu, Z.~Su, C.~Pang, and X.~Luo, ``Cascading and
  enhanced residual networks for accurate single-image super-resolution,''
  \emph{IEEE Trans. Cybernetics}, pp. 1--11, 2019.

\bibitem{dualattention}
F.~{Li}, H.~{Bai}, and Y.~{Zhao}, ``Learning a deep dual attention network for
  video super-resolution,'' \emph{IEEE Trans. Image Process.}, vol.~29, pp.
  4474--4488, 2020.

\bibitem{SOFVSR}
L.~Wang, Y.~Guo, Z.~Lin, X.~Deng, and W.~An, ``Learning for video
  super-resolution through hr optical flow estimation,'' in \emph{Lect. Notes
  Comput. Sci.}, C.~V. Jawahar, H.~Li, G.~Mori, and K.~Schindler, Eds.\hskip
  1em plus 0.5em minus 0.4em\relax Cham: Springer International Publishing,
  2019, pp. 514--529.

\bibitem{SOFVSR-TIP}
L.~{Wang}, Y.~{Guo}, L.~{Liu}, Z.~{Lin}, X.~{Deng}, and W.~{An}, ``Deep video
  super-resolution using hr optical flow estimation,'' \emph{IEEE Trans. Image
  Process.}, vol.~29, pp. 4323--4336, 2020.

\bibitem{ye2017algorithm}
Y.~Ye, E.~Alshina, and J.~Boyce, ``Algorithm descriptions of projection format
  conversion and video quality metrics in 360lib,'' \emph{Joint Video
  Exploration Team of ITU-T SG}, vol.~16, 2017.

\bibitem{fastnet}
R.~Lan, L.~Sun, Z.~Liu, H.~Lu, C.~Pang, and X.~Luo, ``Madnet: A fast and
  lightweight network for single-image super resolution,'' \emph{IEEE Trans.
  Cybernetics}, pp. 1--11, 2020.

\end{thebibliography}



%







\end{document}